\documentclass[final]{cvpr}

\usepackage{times}
\usepackage{epsfig}
\usepackage{graphicx}
\usepackage{amsmath}
\usepackage{amssymb}

\usepackage[pagebackref=true,breaklinks=true,colorlinks,bookmarks=false]{hyperref}
\usepackage[T1]{fontenc}

\def\confYear{CVPR 2021}

\begin{document}

\title{PAL: Intelligence Augmentation using Egocentric Visual Context Detection}

\author{Mina Khan\\
MIT Media Lab\\
75 Amherst St, Cambridge, MA, USA\\
{\tt\small minakhan01@gmail.com}
\and
Pattie Maes\\
MIT Media Lab\\
75 Amherst St, Cambridge, MA, USA\\
{\tt\small pattie@media.mit.edu}
}

\maketitle

\begin{figure*}[]
\centering
  \includegraphics[width=0.95\linewidth]{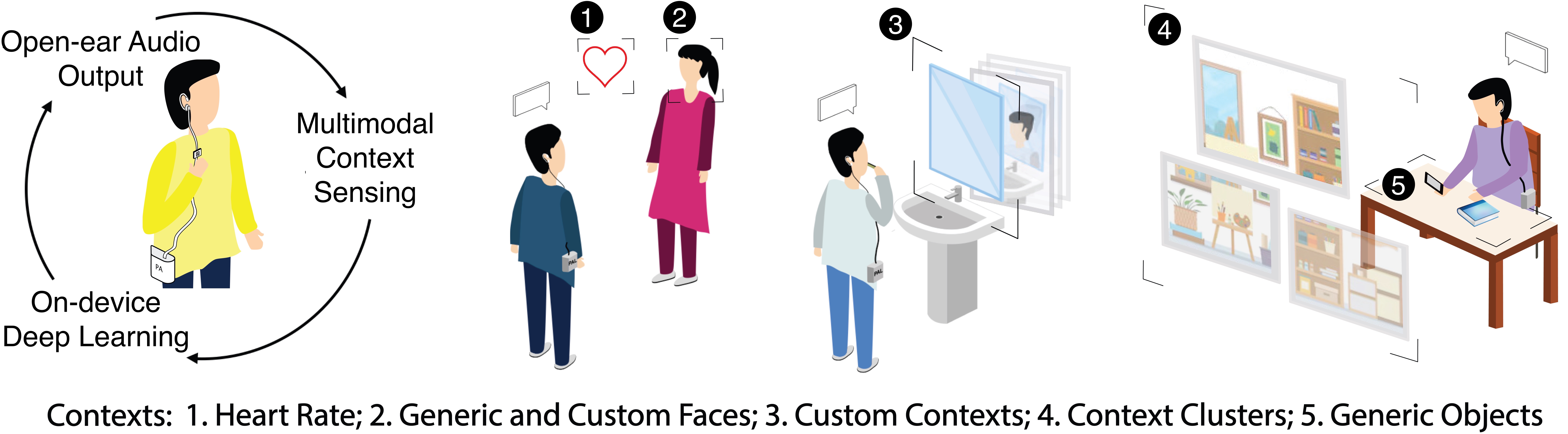}
  \caption{Wearable system using multimodal sensors and on-device deep learning for context-aware support in egocentric visual contexts.}
  \label{fig:teaser}
\end{figure*}

\begin{abstract}
Egocentric visual context detection can support intelligence augmentation applications. We created a wearable system, called PAL, for wearable, personalized, and privacy-preserving egocentric visual context detection. PAL has a wearable device with a camera, heart-rate sensor, on-device deep learning, and audio input/output. PAL also has a mobile/web application for personalized context labeling. We used on-device deep learning models for generic object and face detection, low-shot custom face and context recognition (e.g., activities like brushing teeth), and custom context clustering (e.g., indoor locations). The models had over 80\% accuracy in in-the-wild contexts (\textasciitilde1000 images) and we tested PAL for intelligence augmentation applications like behavior change. We have made PAL is open-source to further support intelligence augmentation using personalized and privacy-preserving egocentric visual contexts.

\end{abstract}

\section{Introduction}
Egocentric visual contexts have been useful in context-aware intelligence augmentation \cite{engelbart1962augmenting}, e.g., memory augmentation and assistive technology \cite{xia_design_2013, doherty_wearable_2013}.
However, visual context tracking, especially sending user data to the cloud for deep learning, raises privacy concerns and deep learning models also have to be personalized for each user.

We created a wearable system, called PAL (Figure \ref{fig:teaser}), for personalized and privacy-preserving egocentric visual context recognition using on-device, human-in-the-loop, and low-shot deep learning.
PAL uses on-device deep learning for real-time and privacy-preserving processing of user data, and includes multimodal sensing, e.g., using camera, heart-rate, physical activity, and geolocation \cite{khan2021behavior}.
PAL also supports user input for human-in-the-loop training of personalized visual contexts. 
We used on-device models for generic object and face detection, personalized and low-shot custom face and recognition, and semi-supervised custom context clustering \cite{khan2021wearable}.
Compared to existing wearable systems, which use at least 100 training images per custom context \cite{lee2018personalized} and do not use privacy-preserving on-device deep learning, PAL's on-device models for low-shot and continual learning use \textasciitilde10 training images per context. Also, PAL uses active learning for context clustering so that the users do not have to explicitly train different contexts.

We make three contributions:
i. a wearable device for privacy-preserving and personalized egocentric visual context detection using on-device and human-in-the-loop deep learning;
ii. a system for recognizing custom contexts, faces, and clusters using low-shot, custom-trainable, and active learning;
iii. real-world applications and evaluations.

\section{Related Work}
PAL is a wearable system for personalized and privacy-preserving egocentric visual context detection using on-device, human-in-the-loop, and low-shot deep learning.
PAL also includes multimodal sensors and input/outputs.
The existing wearable systems for egocentric visual context detection do not support multimodal sensing and on-device, human-in-the-loop, and low-shot deep learning like PAL.

\subsection{Wearable Cameras and Deep Learning}
Wearable cameras are commonly used for intelligence augmentation applications \cite{xia_design_2013, doherty_wearable_2013} and even combined with physiological sensors \cite{healey1998startlecam}.
Deep learning has also been used with egocentric cameras, e.g., for predicting daily activities \cite{castro_predicting_2015}, eating recognition \cite{bedri2020fitbyte}, visual assistance \cite{mulfari_tensorflow-based_2018, nishajith2018smart}, visual guides \cite{seidenari_deep_2017}, and face recognition \cite{daescu_deep_2019}).
However, none use on-device deep learning, especially for personalized and privacy-preserving egocentric visual contexts.
On-device deep learning systems have also been used for computer vision \cite{mathur2017deepeye}, but they do not support personalized, low-shot, and human-in-the-loop visual context detection.

\subsection{Privacy-preserving Deep Learning}
Privacy-preserving approaches include privacy-preserving collaborative deep learning for human activity recognition \cite{lyu_privacy-preserving_2017} and image distort or modification \cite{dimiccoli2018mitigating, alharbi2019mask}.
However, none of these systems use on-device deep learning to avoid sending data to the cloud for processing. 

\subsection{Personalized Deep Learning}
Wearable deep learning-based egocentric visual context recognition has been used for personalized object and face recognition \cite{lee2016personalized, lee2018personalized}.
However, they do not use on-device deep learning for privacy-preserving context detection and also need \textasciitilde100 images for training each personalized class, whereas our custom context and face recognition models use 1 to 10 images.
There are also approaches for low-shot and continual deep learning \cite{qi2018low} and active deep learning \cite{hossain_active_2019}, including for image clustering \cite{coletta_combining_2019}.
However, none have been deployed in in-the-wild wearable settings, especially for on-device and personalized deep learning.

\section{Design and Implementation}
We created a wearable system, called PAL, with multi-modal sensing (camera, geolocation, heart-rate, and physical activity), on-device deep learning, and audio input/output. 
PAL's wearable device, on-device deep learning models, and mobile/web app are described below.

We made 3 design decisions to enable personalized and privacy-preserving egocentric visual context detection. First, we decided to use a head-mounted egocentric camera to capture the user's visual context, but decided to keep the camera small, light, and non-conspicuous. Second, we decided to use on-device deep learning for model training and inference so that user images were not sent to the cloud for processing and can be processed on-device for privacy-preserving sensing, e.g., hiding bystander's faces. Third, we decided to enable human-in-the-loop and active learning to learn personalized contexts for intelligence augmentation.

\subsection{Wearable Device}
\begin{figure}[]
  \includegraphics[width=\linewidth]{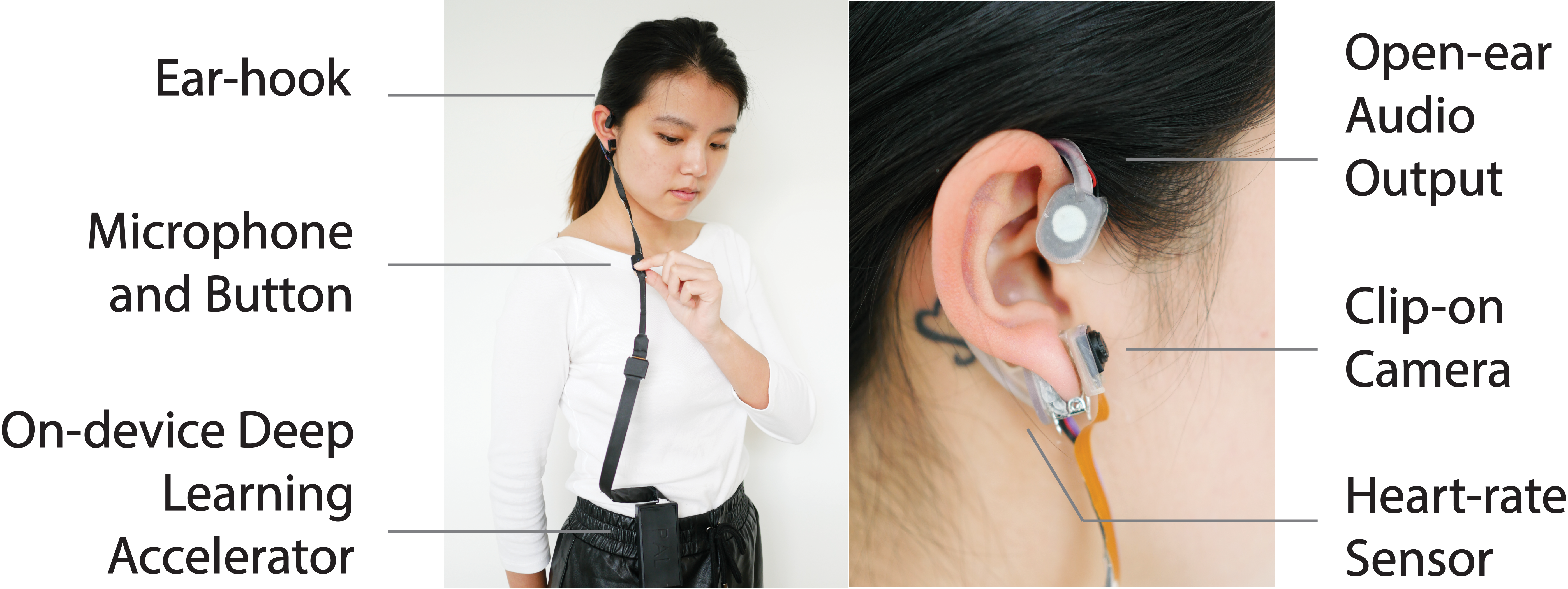}
  \caption{Wearable device with on-device deep learning, camera, open-ear audio output, microphone, and heart-rate sensor.}
  \label{fig:wearable_device}
\end{figure}

PAL’s wearable device has an ear-hook and an on-body component (Figure \ref{fig:wearable_device}). 
The ear-hook has a camera, heart-rate sensor, and open-ear audio output. The on-body component has a microprocessor and a deep learning accelerator.
The device takes periodic pictures, has a 5-hour battery life (0.3A at 5.25V, 2500mAh battery). The device supports audio input/output and the users can press a button for starting/stopping custom face and context training sessions. 

\subsection{On-device Deep Learning Models}
We added on-device visual context detection models for three tasks: i. generic object and face detection; ii. low-shot custom face and context recognition; iii. context clustering for efficient active learning of visually-similar contexts.
All models are trained and inferred on the wearable device.

\textit{Object and Face Detection:} The \emph{Object Detection} model is trained on 90-item Common Objects in Context (COCO) dataset \cite{lin2014microsoft} and \emph{Face Detection} on Open Images v4. Both models use the MobileNet SSD v2 architecture \cite{Howard2017-dn, liu2016ssd}.  

\textit{Low-shot Custom Face and Context Recognition:}
We use MobileNetv1 architectures \cite{Howard2017-dn} for both. \emph{Custom Face Recognition} models using FaceNet \cite{schroff2015facenet} and needs 1-2 custom training images per face. \emph{Custom Context Recognition} model recognizes custom activities, e.g., brushing teeth, using weight imprinting \cite{qi2018low}.
Weight imprinting adds new classes to the existing list of classes for continual learning and needs \textasciitilde10 training images per class.
We used weight imprinting on embeddings from an Image Classification model pre-trained on the 1000-class ImageNet dataset \cite{deng2009imagenet}. 

\textit{Context Clustering for Active Learning:}
We cluster visually-similar contexts, separated by geolocations, using an image embedding generator combined with clustering.
We generated embeddings using an Image Classification model (MobileNet v1), pre-trained on 1000-class ImageNet dataset.
We use Density-based Spatial Clustering of Applications with Noise (DBSCAN) \cite{ester1996density} for clustering as DBSCAN does not require a predetermined number of clusters.

\subsection{Mobile/Web Application}
\begin{figure}[]
  \centering
  \includegraphics[width=0.95\linewidth]{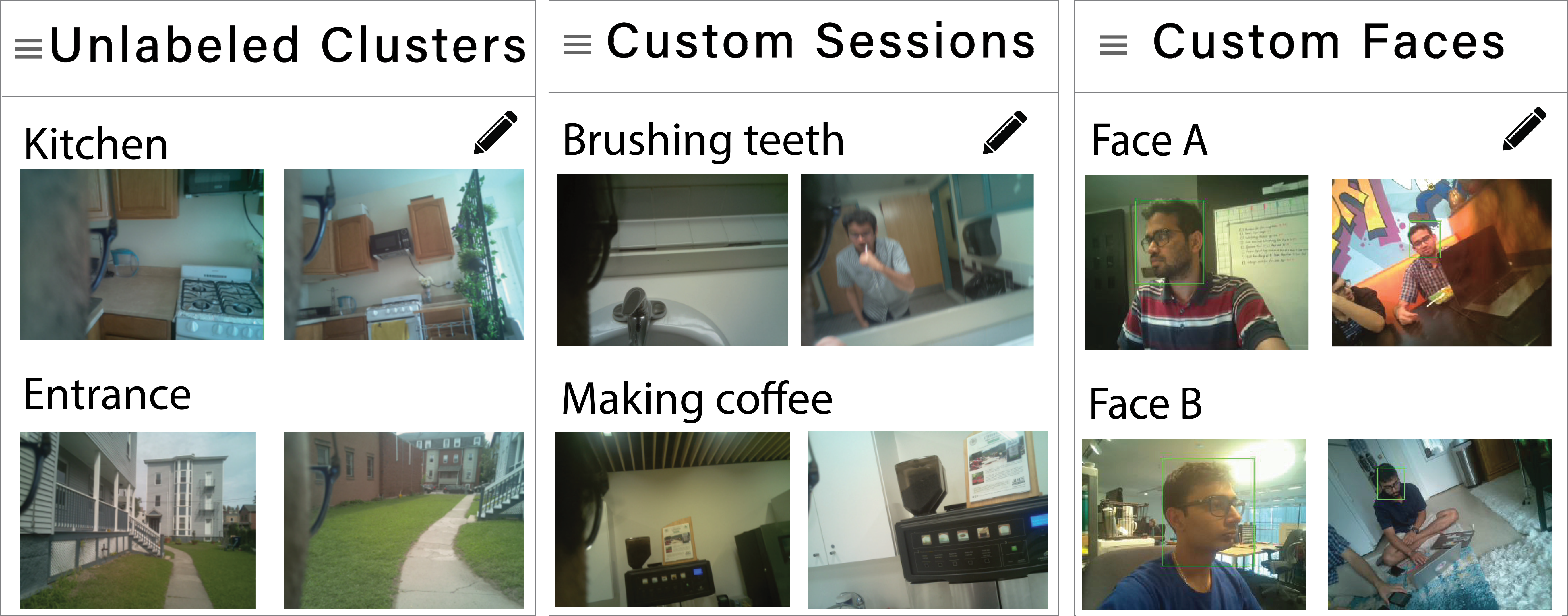}
  \caption{Interface for custom cluster, context, and face labeling.}
  \label{fig:web_app}
\end{figure}

The mobile/web application supports geolocation and physical activity tracking, personalized data labeling and visualization, and custom intervention-setting, e.g., just-in-time reminders for habit-formation support \cite{khan2021improving}.
Custom contexts, context clusters, and faces are shown on the mobile app for labeling. Figure \ref{fig:web_app} shows the labeling interface. 

\section{Evaluations, Applications, and Future Work}
We performed technical evaluations of PAL. We tested PAL's deep learning models with 4 participants for 2 days each (\textasciitilde1000 in-the-wild images) \cite{khan2021wearable}.
Each model had over 80\% accuracy -- Object Detection: 98.8\% (F1 = 0.79, \textasciitilde1000 instances); 
Face Detection: 88.8\% (F1 = 0.9, \textasciitilde180 instances);
Custom Face Recognition: 86.9\% (4 faces, 120 instances); Custom Contexts: 87.2\% (7 activities, \textasciitilde{}350 images);
Custom Clusters: 82\% (19 contexts, \textasciitilde{}300 images). 
Even images partially occluded by cheeks or hair were accurately predicted.
Inference time was \textasciitilde3 sec per model and \textasciitilde15 sec for 5 models.
Moreover, the camera captured \textasciitilde70\% of the user's visual contexts and the heart-rate sensor accuracy was also comparable to state-of-the-art sensors \cite{khan2021behavior}.

We used PAL for real-world habit habit-formation support and showed that habit support in egocentric visual contexts can double habit-formation compared to state-of-the-art techniques \cite{khan2021improving, khan2021pal}. We also built applications for language learning and memory support using PAL \cite{khan2019pal}. 

Users want behavior change support in multimodal contexts \cite{khan2021users}, and we envision that PAL can be used for context-aware behavior change support. PAL's egocentric visual context detection can also be deployed for other intelligence augmentation applications, e.g., memory support for Alzheimer's patients and caretakers. More sensors can be also added to PAL's open-source platform and further work on on-device deep learning will enable better context detection using PAL, e.g., audio-visual context detection.

\section{Conclusion}
Context-awareness is key for intelligence augmentation and egocentric visual contexts can provide rich information about a user's environmental and behavioral contexts. We created a wearable system, called PAL, for personalized and privacy-preserving egocentric visual context detection. PAL uses on-device deep learning for privacy-preserving processing, and we also added multimodal sensing (heart-rate, geolocation, and physical activity) and seamless audio input/output to PAL. 
We deployed on-device, low-shot, and human-in-the-loop deep learning models on PAL to recognize generic faces and objects as well as custom faces, contexts (e.g., activities like brushing teeth), and context clusters (e.g. indoor locations).
We tested PAL in in-the-wild wearable settings and also used it for real-world language learning, behavior change, and memory augmentation applications.
Thus, using on-device, low-shot, and human-in-the-loop deep learning, PAL paves the way for intelligence augmentation applications using personalized and privacy-preserving egocentric visual contexts.

{\small
\bibliographystyle{ieee_fullname}
\bibliography{egbib}
}

\end{document}